\documentclass[10pt,twocolumn,letterpaper]{article}

\usepackage{cvpr}
\usepackage{times}
\usepackage{epsfig}
\usepackage{graphicx}
\usepackage{amsmath}
\usepackage{amssymb}
\usepackage{multirow}
\usepackage{floatrow}
\usepackage{tabularx}
\usepackage{array}
\newfloatcommand{capbtabbox}{table}[][\Fbwidth]
\usepackage{booktabs}
\usepackage{arydshln} 
\usepackage{subcaption}

\usepackage[breaklinks=true,bookmarks=false]{hyperref}

\cvprfinalcopy 

\def\cvprPaperID{****} 

\begin{document}

\title{Mode Penalty Generative Adversarial Network with adapted Auto-encoder}

\author{Gahye Lee, Seungkyu Lee \\
The college of Electronics and Information, Kyung hee University, Republic of Korea\\
{\tt\small \{waldstein94, seungkyu\}@khu.ac.kr}}

\maketitle

\begin{abstract}
Generative Adversarial Networks (GAN) are trained to generate sample images of interest distribution. To this end, generator network of GAN learns implicit distribution of real data set from the classification with candidate generated samples.
Recently, various GANs have suggested novel ideas for stable optimizing of its networks. However, in real implementation, sometimes they still represent a only narrow part of true distribution or fail to converge. We assume this ill posed problem comes from poor gradient from objective function of discriminator, which easily trap the generator in a bad situation.
To address this problem, we propose a mode-penalty GAN combined with pre-trained auto-encoder for explicit representation of generated and real data samples in the encoded space. In this space, we make a generator manifold to follow a real manifold by finding entire modes of target distribution.
In addition, penalty for uncovered modes of target distribution is given to the generator which encourages it to find overall target distribution.
We demonstrate that applying the proposed method to GANs helps generator's optimization becoming more stable and having faster convergence through experimental evaluations.
\end{abstract}
Recently, generative networks have been widely studied thanks to the explosive and successful applications of Generative Adversarial Networks (GAN) proposed by \cite{goodfellow2014generative}.
\begin{figure}[!t] 
  \centering
  \includegraphics[width=0.8\linewidth]{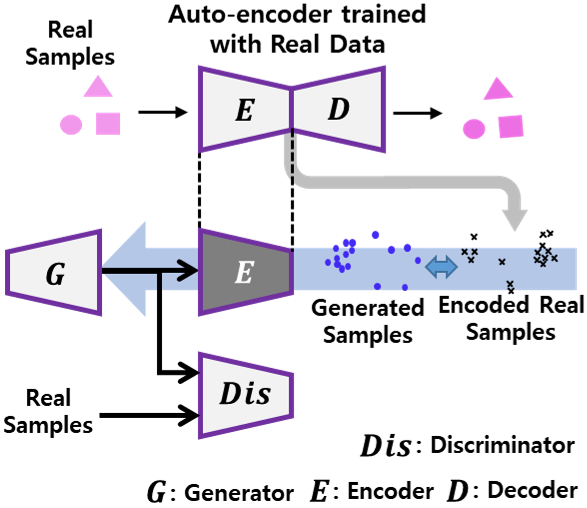}
  \caption{Proposed Generative Adversarial Network with Auto-encoder: Pre-trained real data encoder enables explicit similarity measurement of two distributions in the encoded low-dimensional space.} 
  \label{processing}
\end{figure} 
GAN is composed of two main networks: discriminator and generator.
The core idea is learning the target distribution through two-player minimax game theory. The discriminator helps the generator learn the representation of the target distribution by distinguishing the difference between generated and real samples. As a result, the generator is able to produce samples which resemble target real data. 
Although GAN has been successfully implemented in many applications, it suffers from ill-training problems such as oscillation between modes, namely mode collapse problem. In \cite{arjov2017towards}, authors show that there is always a perfect discriminator being able to distinguish $P_r$ and $P_g$ when the supports of real and generator distribution are disjointed in low dimensional manifolds. Because the gradient of generator is vanishing from the better discriminator, the discriminator is not able to make generator learn anything by back-propagating.

In such condition, once the generator finds a local optimal point of the discriminator in each generator training step, the network starts to produce same samples that minimizes the expectation value in objective function regardless of input noise vectors.
Consequently, generator repeatedly generates certain good instance rather than diverse instances from entire area of real data distribution.

Motivated from this, we propose mode penalty GAN combined with auto-encoder for explicit representation of generated and real data samples in the encoded space. 
Our basic idea is to let the generator know the set of target modes during learning. If there is a set of modes well-representing a typical distribution, it is helpful for training the generator to force the generated samples to follow a set of modes in the target distribution. 
We believe that auto-encoder trained with target data set has an ability to represent the distribution of target data set in the latent space under optimally minimized dimensions.
Therefore, the encoded space representation allows us to compare real distribution and the distribution of generated samples in an explicit manner. In this space, when produced samples are placed near a few target modes, known as falling in mode collapse problem, the penalty for uncovered modes is given to the generator and it encourages the generator in finding an overall target distribution.  
To this end, we expand generator objective function to save the diversity of target distribution via adopted auto-encoder which works as a connecting bridge between original data and manifold space. 
As a result, proposed mode-penalty GAN not only resolves mode collapse problem, but also improves the convergence on target distribution. 

\section{Related Work}
BiGAN \cite{donahue2016adversarial} proposes unsupervised feature learning framework adding encoder as an inverse of generator to represent feature space of data.
Unrolled GAN \cite{metz2016unrolled} proposes to train generator using unrolled multiple samples rather than a single target at the generator output, which alleviates mode collapsing.
VEEGAN \cite{srivastava2017veegan} employs a reconstruction network which not only maps real data distribution to a Gaussian in encoded space, but also approximates reverse action of the generator. As VEEGAN estimates implicit probability of real data set, it prevents mode collapsing problem and produces more realistic and diverse samples. 
\cite{bang2018mggan} proposes a manifold guided generative adversarial network (MGGAN) for guiding a generator by adding another adversarial loss in the manifold space transformed by pre-trained encoder network. This enables the generator to learn all modes of target distribution without impairing the quality of an image. Although real data distribution is represented in the manifold loss of their second discriminator, it is not able to estimate explicit distance between the distributions. 
OT-GAN(\cite{salimans2018improving}) expands generator loss function using optimal transport theory. They define a new metric, Mini batch Energy Distance, over probability distribution in a adversarially learned feature space. Although they make the condition of GAN more stable than other state-of-the-art methods, the computational cost of their large mini-batches is too expensive to be practical in real applications. 
\cite{mescheder2017numerics} identifies reasons of mode collapse problem as both the presence of eigenvalues of the Jacobian of the gradient vector and eigenvalues with big imaginary part. It proposes dynamics-based approaches for avoiding mode collapse showing local convergence to Nash-equilibrum.
Recently, PacGAN \cite{lin2017pacgan} is introduced that simply extends discriminator input for packed with multiple samples showing state-of-the-art mode detection performance. 


In all precedent approaches, using the increased number of samples or larger mini-batches for the training of generator network allows more stable and improved performance. On the other hand, they require hugely increased computation cost and training time.  
Our idea is that pre-trained auto-encoder with real data set provides encoded space representation for both real and generated samples, which allows us to measure the distance between two distributions. 
Given the limited number of generated samples in the training, representation ability of the samples can be evaluated and maximized for both distributions (current generator outputs and real data set) adopting mode penalty weight estimation of the samples.


\section{Mode-Penalty GANs}
\subsection{Generative Adversarial Networks}
Generative Adversarial Network (GAN) is motivated by game theory in which two players (generator and discriminator) compete with each other in a zero-sum game framework. The generator learns the target distribution via an adversarial training process with the discriminator. When the generator transforms a noise vector $z$ into a data vector $G(z)$ on the target space, the discriminator tries to estimate the probability if input sample comes from target distribution $P(x)$ or not. At the end of the training process, the generator produces outputs of $P(x)$. GAN defines this adversarial problem as follows:

\[
\min_{G}\max_{D}L_{GAN}(D,G) = \mathbb{E}_{x\sim P_{data}}\log{D(x)}
\]
\begin{equation}\label{GAN loss}
    + \mathbb{E}_{z\sim P_z}\log(1-D(G(z)))
\end{equation}
The goal of GAN is to find the optimal parameter set of $G$ and $D$ in equation (\ref{GAN loss}) minimizing the generator loss and maximizing the discriminator loss. 
  
\subsection{Distance between Two Manifolds} 
Assume that there is a set of optimal mode samples in manifold space of real data and also a set of samples from generator composing modes of generator's manifold space. Our hypothesis is that forcing modes of generator to follow modes of real distribution encourages the minimization the distance between two manifolds and finally leads to help the learning of GANs. 
Real data distribution cannot be explicitly represented in original data space and we expect to transform real data to lower dimensional space for better understanding and easier manipulation. 
Auto-encoder is a neural network that learns discriminative representation of unlabeled input data in its learned encoded space. In other words, it learns compact and abstracted low dimensional representation of real data and is able to generate original input through decoding network.
In the encoded space of auto-encoder, abstracted feature set represents aspects of original data distribution. Therefore, we expect that the encoded space provides more tractable way of defining mode and distance calculation between two manifold spaces for the training of our generative network. 

\begin{figure}[!t] 
  \centering
  \includegraphics[width=0.9\linewidth]{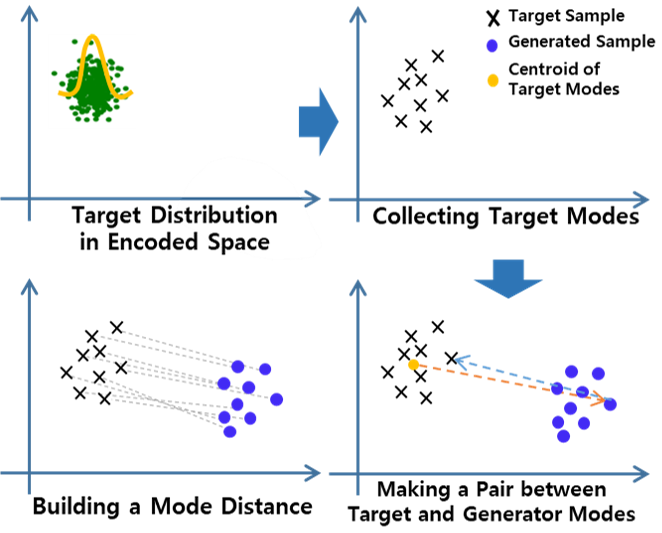}
  \caption{Distance calculation between generated and real data samples in an encoded space} 
  \label{proc}
\end{figure}

There are various way for collecting target modes, for instance, sampling from the target distribution estimated using KDE, or selecting the samples from each class by defining heuristically. Since the target samples considered as modes will be used to evaluate the similarity of generated samples in the encoder space, they have to be guaranteed to be come from entire modes of real data distribution. 
After extracting $N$ target from real modes of target distribution and $N$ generated samples from generator, our network is able to compare real and generated manifold in the encoded space through adding new loss term to generator objective function. 
We construct distance loss between two manifolds through mode matching algorithm.
Taking a MoG dataset as an example, the matching procedure is summarized in Figure \ref{proc}. For every generated sample, we select the furthest generated sample from a centroid of real modes. Then, a real sample which has a closest distance from the selected one is chosen. Sample used in matching once are not reused. After making a generated-real sample pairs, we measure the distance the pair has for entire pair defined in equation (\ref{distance}) namely mode distance loss. 
\begin{equation}\label{distance}
Dist(X_r,X_g) =\mathbb{E}_{x_r:x_g} w_{p}\sqrt{\left \| x_{r}-x_{g} \right \|^2}
\end{equation}   
$X_r$ and $X_g$ denote a set of modes of real and generated distribution. $x_{r}$ and $x_{g}$ respectively denote paired real and generated mode in encoded space. And $w_p$ denotes a weight of penalty for missing modes which will be discussed in more detail in the next subsection. Intuitively, if the generator produces samples near only a few modes of target, then distance from missing modes are increasing. Penalty about this distance is given to the generator and it rescues the generator from bad situation when the gradient from the discriminator is almost zero. As a result, this process encourages the generator to follow manifold of real data. Next, we will discuss about 

\subsection{Penalty Weight for Missing Modes} 
For finding complete modes of real data distribution, we focus more on missing modes than already covered mode. 
Model samples of real data obtained in the previous section do not have to maintain an equal importance in the distribution similarity measurement.
Rather than some real samples which have been frequently recovered by the generator samples frequently, other real samples that are seldom recovered have to gain more importance in the following training iterations.  
Therefore, we expect to have a different focus on each samples to drive the generator in having diversity in its generated samples.
For example, if our network observes that some generated samples are repeatedly close to a specific real sample, the real sample gets relieved of penalty weight. On the other hand, similarly generated samples that correspond with other distant real sample should be treated with an penalty weight and incur the generator training, which scatters the similarly generated samples to other neighbour modes. Giving heavier penalty for uncovered modes leads generator to find the uncovered mode much faster. 

The basic idea of penalty weight is the quality evaluation of current real and generated samples and reflection of them in the loss calculation of training iteration.  
Real sample which have already been recovered by generated samples are assigned with lower penalty weight.
Previously, we defined the mode distance as $Dist(X_r,X_g)$ which measures the distance between pairs of real and generated mode in the encoded space of auto-encoder. Now, we adjust the importance of each real samples $x_r$ using penalty weight.

Penalty weight $w_p$ in equation (\ref{modepenalty}) indicates how well all generated samples contribute to the estimation of out model samples of the real data.
We assign this weight for each real sample based on the degree of how much particular real sample is incarnated by generated samples in the past $k$ iterations. Calculated distance of a real sample from paired generated samples in the past iterations adds importance to the real sample. In other words, if a real sample has higher distances to all matched generated samples during the past $k$ training steps, that means this particular real sample is out of the generator interest and it gets higher penalty weight. 
As a result, in the following training steps, the generator is forced to consider this forgotten real sample more with higher loss value.


\begin{equation}\label{modepenalty}
\begin{split}
w_p = \frac{1}{k}\sum_{it=1}^{k}\sqrt{\left \| x_r- x_g^{it} \right \|^2}
\end{split}
\end{equation} 

Finally, penalty weight for each real sample encourages generated samples to cover entire target real data samples. At each training step, this weight is updated to adjust the importance of generated and target samples. 
Due to the elaborate evaluation of the generated sample quality at each training step, proposed network converges to target real distribution faster without a mode collapse problem.
For a set of real modes $X_{r}$ with $x_r$ as the element and a set of generator modes $G(Z)$ with $G(z)$, $z\sim N(0,1)$ as the element, we define the complete generator loss term in equation (\ref{loss}).
If GAN unexpectedly generates samples from single mode of real data set, second term in equation (\ref{loss}) increases and force the network to find other modes. $\lambda_p$ is hyper-parameter adjusting the effect of penalty loss to an adversarial loss. After the generator covers almost all modes, it is turned off. By focusing on the GAN's original purpose, therefore, generator produces much more diverse samples around each mode. 
\begin{equation}\label{loss}
\min_{G} - \mathbb{E}_z[\log{D(G(z))}] + \lambda_{p}Dist(X_{r}, G(Z)) 
\end{equation}  
\begin{figure}[!t] 
  \centering
  \includegraphics[width=1\linewidth]{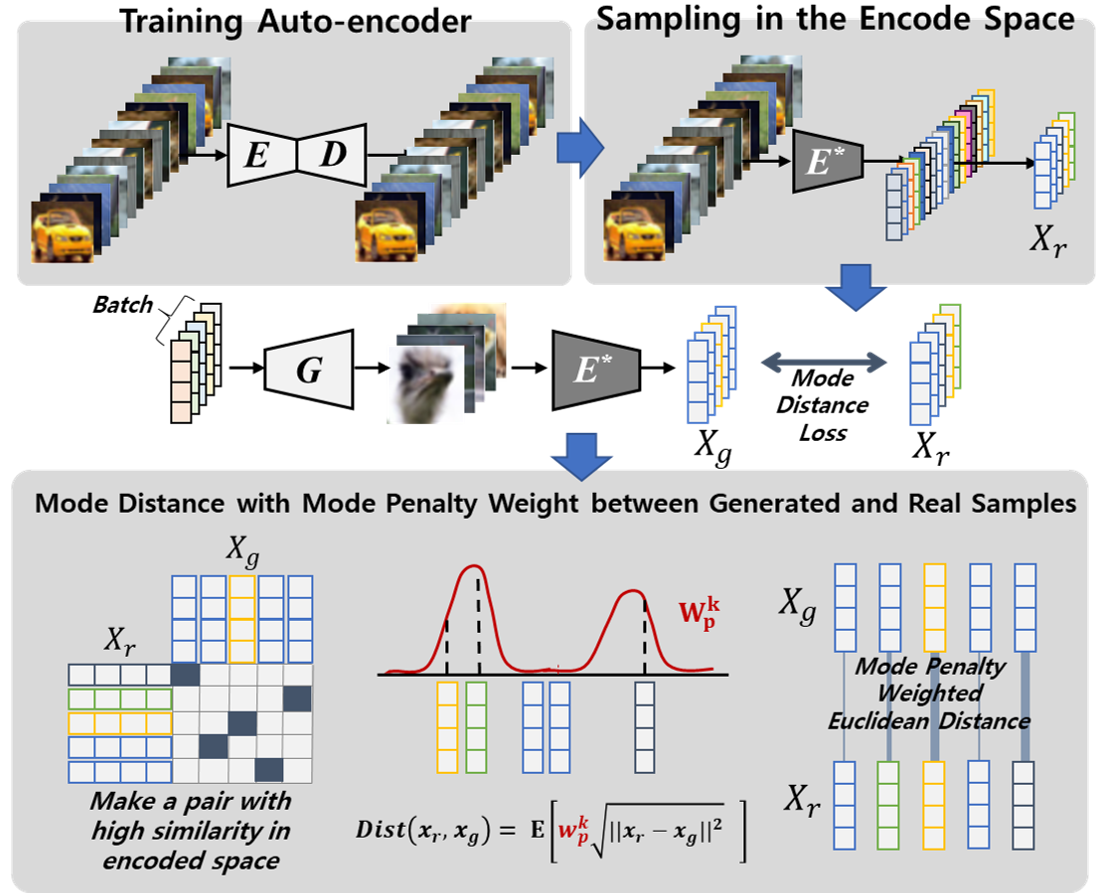}
  \caption{Overall training procedure of generator through adopted auto-encoder. $X_r$ and $X_g$ denote a set of modes in target distribution and generator distribution respectively. $W_p$ is estimated penalty distribution for uncovered mode $k^{th}$ training step. $E$ and $E^*$ denote encoder and pre-trained encoder.} 
  \label{overal_pro}
\end{figure} 

The overall training procedure of generator is summarized in Figure \ref{overal_pro}. First, we train the auto-encoder with target real data set. A set of target modes is represented in encoded space as $X_r$. Note that the real samples are extracted once and used for entire iterations of the training. After producing a set of modes $X_r$ by generator in same space, we make pairs between modes and each pair is given the mode penalty weight according to the distance.   

\begin{figure}[!b]
  \centering
  \includegraphics[width=1\linewidth]{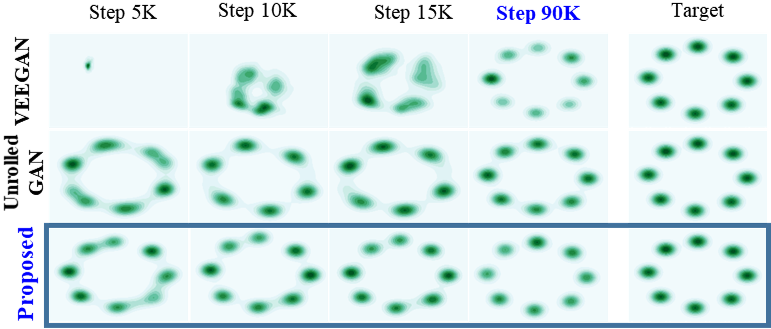}\\
  (a) 2D Ring: 8 Gaussians \\
\includegraphics[width=1\linewidth]{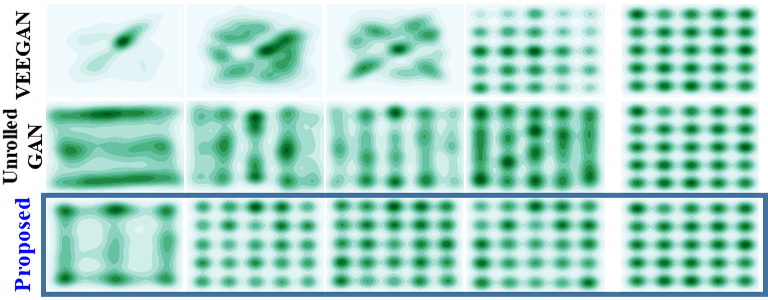}\\
  (b) 2D Grid: 25 Uniform Gaussians \\
\includegraphics[width=1\linewidth]{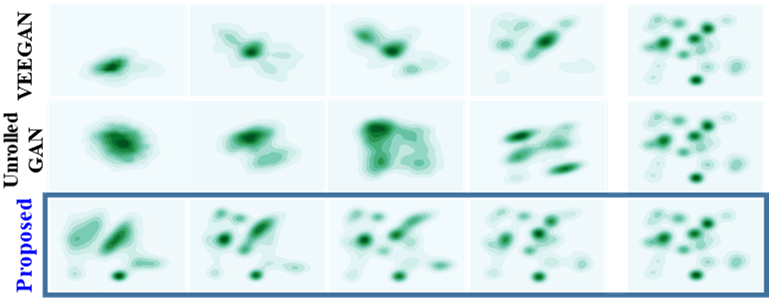}\\
  (c) 2D Random: 25 Random Gaussians \\
\includegraphics[width=1\linewidth]{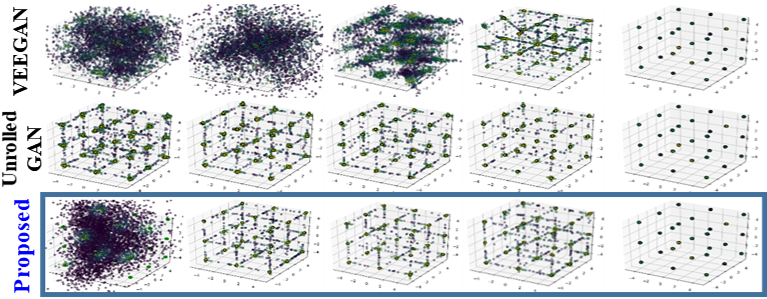}\\
  (d) 3D Cube : 27 Uniform Gaussians
  \caption{Experimental results compared to VEEGAN(\cite{srivastava2017veegan}) and Unrolled GAN(\cite{metz2016unrolled}): 2D Ring, 2D Grid, 2D Random, and 3D Cube. Proposed method converges faster and more accurately than the previous methods.} 
  \label{mogres}
\end{figure}




\section{Experimental Evaluation}
In order to verify the performance of the proposed method, we perform quantitative and qualitative evaluation on three data sets: 1) Four types of synthetic mixture of Gaussians, 2) Stacked MNIST, and 3) Cifar10. We follow quantitative evaluation method of previous methods including additional similarity measure and synthetic data for diverse evaluation.

\subsection{Mixture of Gaussians}
For the analytic and controllable performance evaluation and visualization, 2 dimensional mixture of Gaussian (MoG) distribution data have been used for experimental comparison.
We create four types of MoG distributions: 2D Ring (eight 2D Gaussians located in a ring), 2D Grid (25 2D Gaussians located in a grid), 2D Random (25 2D Gaussians of random amplitude and location), and 3D Cube (27 3D Gaussians in a cube). We create 20,000 real data samples from each MoG model and use for training and use 5000 generated sample for testing.
Experiment is conducted with two-dimensional encoded space. We optimize both networks with the Adam optimizer with the learning rate of 1e-4, we assume number of 500 modes is enough to represent target distribution.   
We choose $\lambda_p$ is 3 which is experimentally giving better outcome. 
For fair comparison, we use identical network implemented in VEEGAN(\cite{srivastava2017veegan}) which is also used in other research.
Generator has three layers of fully connected MLPs with 128 nodes without dropout and batch normalization, and discriminator has two layers of fully connected MLPs without dropout and batch normalization. 
We employ three metrics for quantitative evaluation: number of modes found, high quality sample ratio (HQS) \cite{srivastava2017veegan}, and additional distribution distance measurement Jensen-Shannon divergence (JSD).
Number of Gaussian modes (8, 25, 25, and 27, respectively) found with generated samples and high quality sample ratio (HQS) are counted 20 times and mean and std. are calculated. 

\begin{table*}
\begin{center}
    \caption{Quantitative Evaluation on Mixture of Gaussians: Number of modes found, HQS(High Quality Sample), and JSD(Jensen-Shannon divergence) between real and generated sample distributions. }
\label{eval_table}
\begin{tabular}{cr|rrrrrr} %
\hline
			& METRIC 	&   GAN\cite{goodfellow2014generative} & ALI\cite{dumoulin2016adversarially}&	Unrolled\cite{metz2016unrolled} & VEEGAN\cite{srivastava2017veegan}   & PacGAN4\cite{lin2017pacgan}  & \textbf{Proposed} \\
\hline
			&Modes (Max 8)	    &   6.3 (0.5)   & 6.6 (0.3) &   8.0 (0.0)		& 8.0 (0.0) 	&   7.8(0.1)	        & \textbf{8.0} (0.0)  \\
2D Ring  	&\% HQS	            &   98.2 (0.2)  & 97.6 (0.4)&   30.9 (0.6)		& 60.4 (0.5)    & \textbf{95.9}(0.01)   & 94.0 (0.4)  \\
	        &JSD	    	    &   N/A	        & N/A       &0.25 (0.004)       & 0.19 (0.005)  & N/A	                & \textbf{0.04} (0.005)  \\
\hline
			&Modes (Max 25)	    &   17.3 (0.8)  & 24.1 (0.4)&   25.0 (0.0)		& 24.1 (0.3) 	&24.8(0.2) 	            & \textbf{25.0} (0.0)  \\
2D Grid 	&\% HQS			    &   94.8 (0.7)  & 95.7 (0.6)&   14.4 (0.6)	    & 65.4 (0.6)    &\textbf{93.6}(0.6)     & 93.1 (0.4)  \\
        	&JSD	   	        &   N/A         &   N/A     &   0.47 (0.006)	& 0.21 (0.004)  &   N/A    	            & \textbf{0.05} (0.003)  \\
\hline
    		&Modes (Max 25)	    &   N/A  	    &  N/A      &	24.4 (0.5)		& 24.3 (0.6)    &N/A	                & \textbf{25.0} (0.0)  \\
2D Random 	&\% HQS	            &   N/A  	    &   N/A     &	15.9 (0.5)		& 63.7 (0.7)    &N/A                    & \textbf{83.4} (0.6)  \\
        	&JSD		        &   N/A  	    &   N/A     &	0.38 (0.007)	& 0.32 (0.007)  &N/A	                & \textbf{0.15} (0.004)  \\
\hline
			&Modes (Max 27)     &   N/A 	    &   N/A	    &   27.0 (0.0)	    & 26.6 (0.5)    &N/A                    & \textbf{27.0} (0.0)  \\
3D Cube 	&\% HQS		        &   N/A 	    &  N/A      &  85.0 (0.3)	    & 43.3 (0.7)    &N/A	                &  \textbf{85.4} (0.6)  \\
			&JSD	            & N/A  	        &  N/A      &   0.19 (0.004)	& 0.31 (0.005)  &N/A                    & \textbf{0.12} (0.003)  \\  
\hline
\end{tabular}
\end{center}
\end{table*}

\begin{figure}[!b]
    \centering 
        \includegraphics[width=0.49\linewidth]{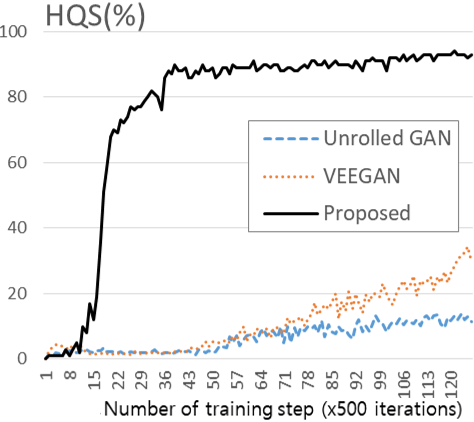}
        \includegraphics[width=0.49\linewidth]{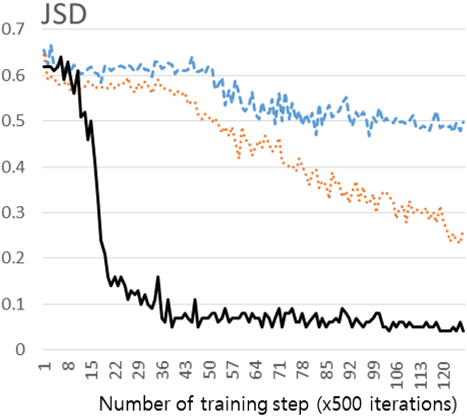} 
    \caption{HQS and JSD of 2D Grid test at each training iteration showing the convergence of compared methods}
    \label{fig:graphs}
\end{figure}

The number of modes found and percentage of high quality samples are not enough to tell how well generated samples are able to cover the entire real data distribution. 
Therefore, we measure the distance between real and generated distributions. We map the samples to canonical space and count the number of points fall in each canonical unit. By measuring Jensen-Shannon divergence (JSD) between the two distributions ($P_{r}$, $P_{g}$), we evaluate how well generator follows target real data (mixture of Gaussians) distribution. 
Figure \ref{mogres} compares both convergence and mode detection performance on the four MoG types.
Proposed method outperforms compared methodss in mode detection performance without mode collapse. For 2D Ring, proposed method already finds all modes within 5K iterations. 
In 2D random case, VEEGAN and Unrolled GAN fails to find modes of non-uniform locations and amplitudes, however our method finds almost identical distribution compared to target truth distribution. 
Proposed method also converges and finds all modes much faster than others.
Figure \ref{fig:graphs} plots HQS and JSD of three methods at each training iteration in 2D Grid test. Proposed method shows outstanding convergence in both measurements.
Recently, a new approach called PacGAN is proposed \cite{lin2017pacgan} that modifies the discriminator to make decisions based on multiple samples from the same class. This simple idea is tested on existing GANs showing outstanding performance in addressing mode collapse problem. 
We summarize quantitative evaluation in table \ref{eval_table} and our proposed method still finds slightly more modes on synthetic mixture of Gaussians data sets than all types of PacGANs even though it shows slightly lower HQS (Note that, however, std of proposed method is smaller than PacGAN). 

\begin{figure}[!b]
    \centering
        \includegraphics[width=0.49\linewidth]{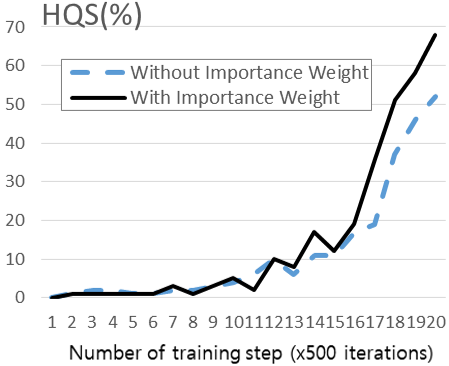}
        \includegraphics[width=0.49\linewidth]{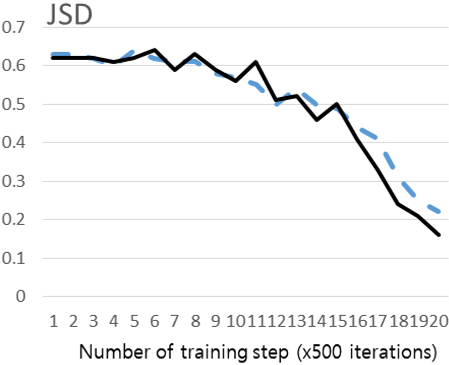}  
    \caption{HQS and JSD of 2D Grid test of our method at each training iteration with and without mode penalty weights, $w_p$, in equation (\ref{modepenalty}) : two measurements show that mode penalty weight encourages faster convergence in the training to all modes}
    \label{fig:wtest}
\end{figure}

Figure \ref{fig:wtest} plots HQS and JSD of 2D Grid test of our method at each training iteration with and without mode penalty weights $w_p$ and  in equation (\ref{modepenalty}). Importance weight achieves around 20\% of HQS gain reaching almost 70\% within 20 epochs even though it shows fluctuating values in very early iterations.

\subsection{Stacked MNIST}   
\label{stackedsection}
In this experiment, we expand MoG dataset to image space with MNIST dataset, and stacked MNIST. 
Evaluation on MNIST data (Figure 4) shows that proposed method generates more number of digits than original GAN.
Stacked MNIST is designed for high complexity evaluation with extended number of modes. Stacked MNIST is synthesized by stacking different MNIST digits in respective colors. This synthesized dataset has 1000 modes which are the combinations of 10 classes in 3 channel. We use implemented discriminator and generator architecture of standard DCGAN. 
Auto-encoder has three convolutional layers with 5 by 5 filters, 2 fully connected layers for encoder part and one fully connected layer, 3 transposed convolutional layers are used for decoder. First, we train the classifier using MNIST dataset which assigns the mode to generated images. In this evaluation, we use enough generated samples and they are given to pre-trained classifier to give them a mode. For example, one sample has three channels which represent different digits. Digit for each sample in the channel is determined by MNIST classifier. 
Experiment is conducted with 50 dimensional encoded space. Learning rate with Adam optimizer for both network is 2e-4, the number of modes, training sample and test generated sample is 1,000, 500,000, 50,000, respectively. We experimentally choose $\lambda_p$ = 500. 


\begin{figure}[!t]
  \centering
  \includegraphics[width=1\linewidth]{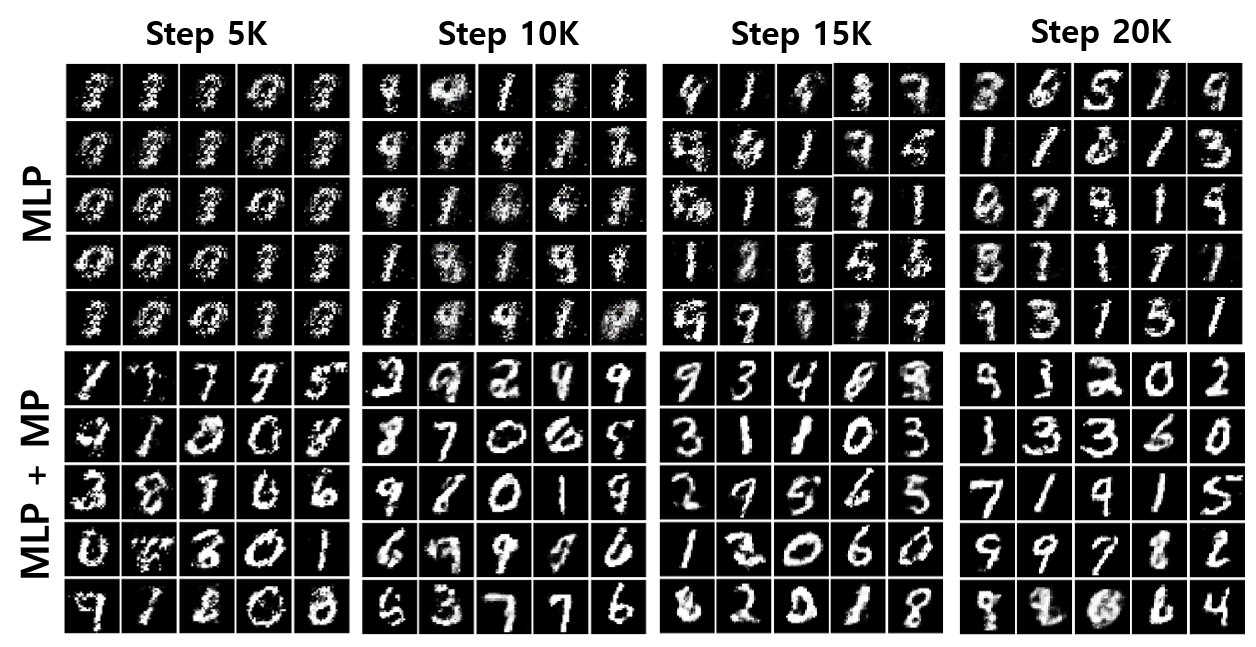}
  \caption{Experimental results on MNIST dataset. First row is original GAN composed mlp and second row is proposed method adding mode penalty loss to mlp GAN.} 
  \label{processing}
\end{figure}


\begin{figure}[!b]
    \centering 
        \includegraphics[width=0.49\linewidth]{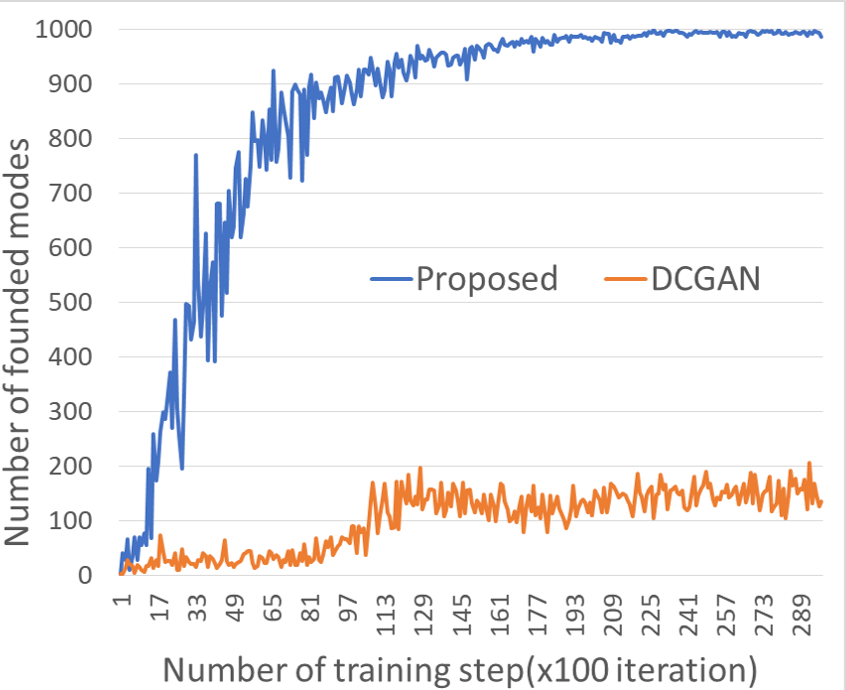}
        \includegraphics[width=0.49\linewidth]{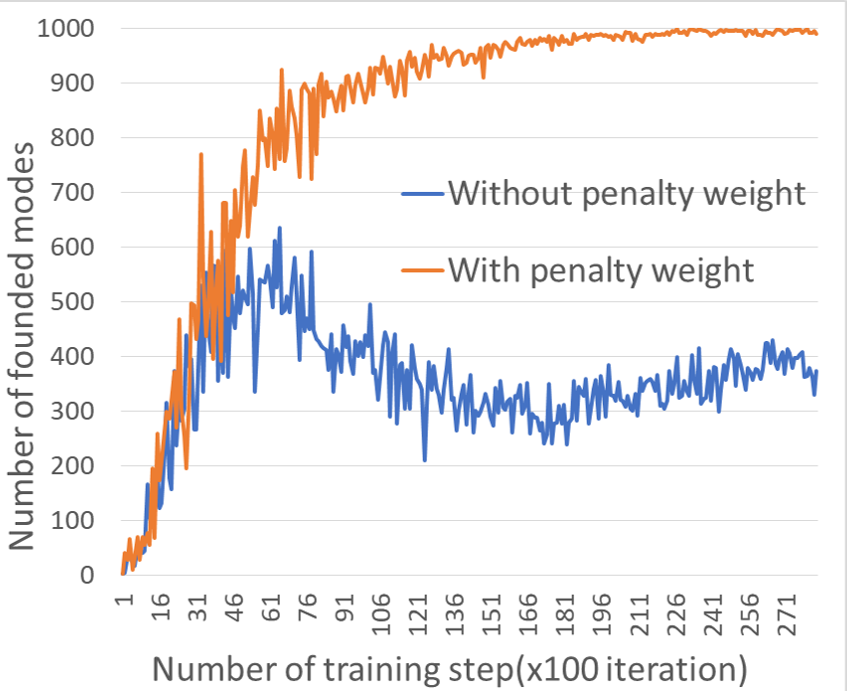} 
    \caption{Finding number of modes on Stacked MNISTs test at each training iteration. Performance on proposed method and DCGAN is recorded in the left chart. The right chart shows the effect of mode penalty weight through weight on/off.}
    \label{fig:eval_stackedmnist}
\end{figure}

\begin{figure}[!t]
    \centering
        \includegraphics[width=0.49\linewidth]{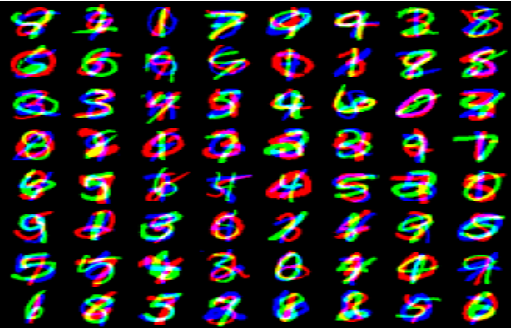}
        \includegraphics[width=0.49\linewidth]{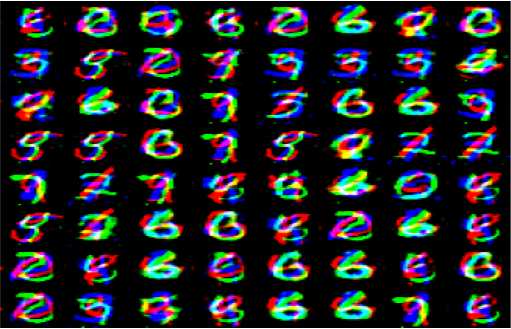}\\
        Target Real \quad \quad\quad\quad\quad\quad \quad DCGAN\\
        \includegraphics[width=0.49\linewidth]{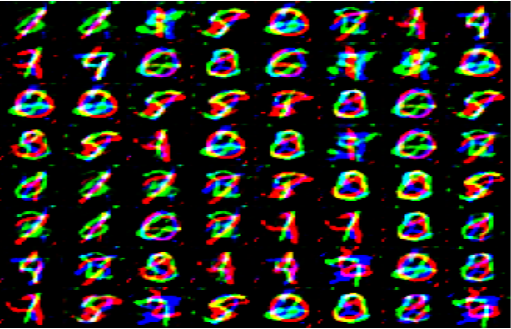}
        \includegraphics[width=0.49\linewidth]{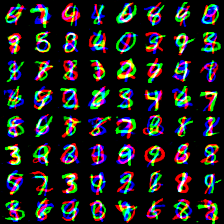}\\
        VEEGAN\quad \quad\quad\quad\quad\quad \quad \textbf{Proposed}
    \caption{Generated stacked MNIST samples from trained models. We refer these images from \cite{srivastava2017veegan} paper except our result.}
    \label{fig:three graphs}
\end{figure}

\begin{table}[t]
\begin{center}
    \caption{Quantitative Evaluation on stacked MNIST: Number of modes found and KL divergence are compared.}
\label{eval_stack_table}
\begin{tabular}{r|cc} %
\hline
			& Modes (Max 1000) &	KL     \\
\hline
ALI\cite{dumoulin2016adversarially}  	        &   16.0   &    5.40     \\
Unrolled GAN\cite{metz2016unrolled}    &   48.7    &   4.32    \\
VEEGAN\cite{srivastava2017veegan}          &   150.0    &  2.95    \\
DCGAN\cite{radford2015unsupervised}	        &   99.0    &   3.40    \\
WGAN\cite{arjovsky2017wasserstein}            &   314.3 (38.54)  &  2.44 (0.170) \\
PacWGAN4\cite{lin2017pacgan}	    &   965.7 (19.07)    & 0.42 (0.094)\\
PacDCGAN4\cite{lin2017pacgan}	    &   1000 (00.00)   &     0.07 (0.005) \\
\textbf{Proposed}		&  \textbf{999.2}(0.79)     &   \textbf{0.19}(0.002) \\ 
\hline
\end{tabular}
\end{center}
\end{table}

We demonstrate proposed method encourages cover up all modes in target distribution through qualitative and quantitative evaluation. In figure \ref{fig:eval_stackedmnist}, each vertical and horizontal axis indicates the number of founded mode out of 1000 stacked modes and the number of training steps respectively. We compare DCGAN with proposed network in first chart. The number of Mode founded by DCGAN is less than 200 and quality of generated sample is also not good as shown in figure \ref{fig:three graphs}. However, the first chart shows our method gives a great support to DCGAN to find uncovered modes and finally it covers all modes in target distribution. We also verified how much mode penalty weight affects finding mode through switching weight on/off. As shown in the second chart, training generator with penalty weight shows greater performance than without it. Based on the result, we confirm that focusing on discovering missing mode ensures fast convergence to covering entire distribution.
Table \ref{eval_stack_table} shows quantitative evaluation on stacked MNISTs with other GANs. We count the modes of generated sample and calculate KL divergence between all target modes and generated modes. Mostly, our method has a superior performance among the others except PacGAN. 
To give further explanation on PacGAN, as packing is a simple embedding technique on existing method, PacGAN uses WGAN and DCGAN for stacked MNIST evaluation (table \ref{eval_stack_table}). 
Without packing embedding, proposed method still shows better performance than all types of PacWGAN and competitive results over PacDCGAN.
Proposed method has shown outperforming evaluation results on existing methods and embedding packing on proposed method may produce improved results. 
\subsection{CIFAR 10}
We also qualitatively evaluate our method using CIFAR-10 dataset, which includes 32x32 color images with 10 classes collected by \cite{krizhevsky2009learning}. The architecture and hyper parameter is same to \ref{stackedsection}. Generated results of VEEGAN and DCGAN are collected from \cite{srivastava2017veegan}.
As shown in Figure 6, VEEGAN and DCGAN frequently include identical samples suffering from mode collapsing (see yellow marks).

\begin{figure}[!]
  \centering
        \includegraphics[width=0.32\linewidth]{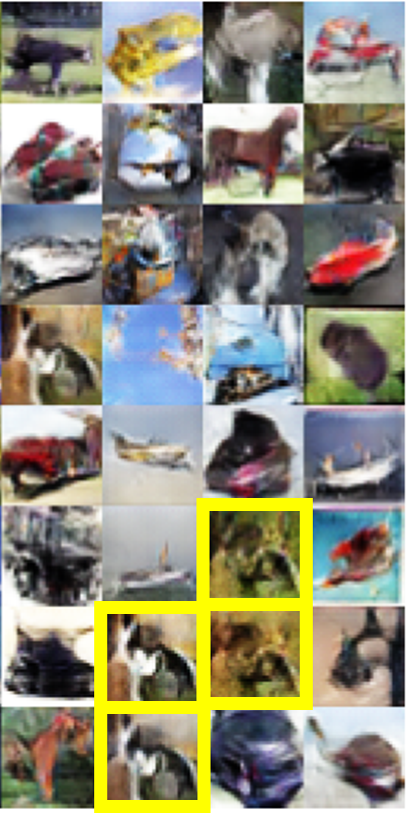}
        \includegraphics[width=0.33\linewidth]{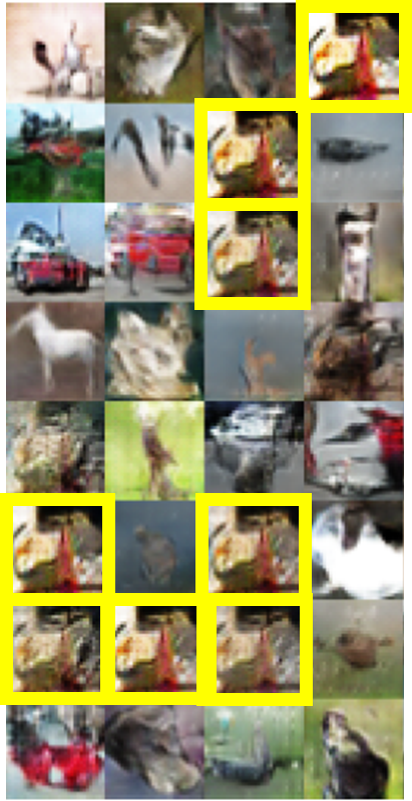}
         \includegraphics[width=0.32\linewidth]{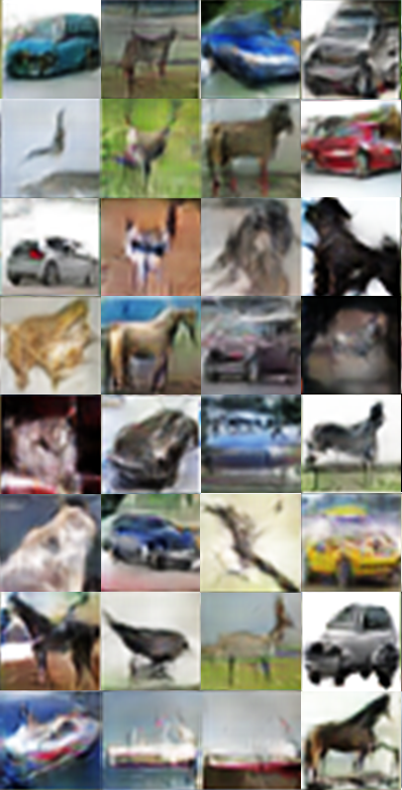}\\
         VEEGAN \quad\quad\quad\quad DCCAN \quad\quad\quad \quad\textbf{Proposed}
  \caption{Experimental results are compared to VEEGAN and DCGAN using CIFAR 10 dataset: VEEGAN and DCGAN frequently suffer from mode collapsing, shown by yellow boxes.} 
  \label{processing}
\end{figure}

\section{Discussion and Conclusion}
In this paper, we propose a mode penalty GANs combined with auto-encoder and mode penalty weights for the candidate samples represented in the latent space of an auto-encoder. Mode distance guides the baseline GANs to reach the global optimum with mode penalty weights, which enhances the represented performance of generated samples for target distribution.
Evaluation and comparison are extensively performed on synthetic and real data sets showing promising performance of the proposed method.

\subsection{Auto-encoder and Processing time}
Proposed method requires pre-trained auto-encoder of proper structure for target real data set.
This involves additional computation cost and processing time compared to other methods.
If we have a pre-trained general auto-encoder with color image set, we can simply utilize it for other applications using color real data set with or without fine tuning step. 


{\small
\bibliographystyle{ieee_fullname}
\bibliography{egbib}
}

\end{document}